\documentclass[runningheads]{llncs}

\usepackage{eccv}

\usepackage{eccvabbrv}

\usepackage{graphicx}
\usepackage{booktabs}
\usepackage{algorithm} %
\usepackage{algorithmic} %
\usepackage{array}
\usepackage{color ,xcolor}
\usepackage{colortbl}
\usepackage{subcaption}

\definecolor{shapecolor}{rgb}{0.0,0.5,0.0}

\usepackage[accsupp]{axessibility}  

\usepackage{hyperref}

\usepackage{orcidlink}

\begin{document}

\title{Famba-V:
Fast Vision Mamba with Cross-Layer Token Fusion
} 

\titlerunning{Famba-V}

\author{Hui Shen\orcidlink{0009-0005-1777-9027} \and
Zhongwei Wan\orcidlink{0000-0002-2790-0290} \and
Xin Wang\orcidlink{0009-0007-6483-9357} \and
Mi Zhang\orcidlink{0000-0001-7002-6757}}

\authorrunning{H.~Shen et al.}

\institute{The Ohio State University\\
\email{\{shen.1780,wan.512,wang.15980,mizhang.1\}@osu.edu}\\
\href{https://github.com/AIoT-MLSys-Lab/Famba-V}{https://github.com/AIoT-MLSys-Lab/Famba-V}}

\maketitle

\begin{abstract}
Mamba and Vision Mamba (Vim) models have shown their potential as an alternative to methods based on Transformer architecture.
This work introduces \textbf{F}ast M\textbf{amba} for \textbf{V}ision (\textbf{Famba-V}), a cross-layer token fusion technique to enhance the training efficiency of Vim models. 
The key idea of Famba-V is to identify and fuse similar tokens across different Vim layers based on a suit of cross-layer strategies instead of simply applying token fusion uniformly across all the layers that existing works propose.
We evaluate the performance of Famba-V on CIFAR-100.
Our results show that Famba-V is able to enhance the training efficiency of Vim models by reducing both training time and peak memory usage during training. Moreover, the proposed cross-layer strategies allow Famba-V to deliver superior accuracy-efficiency trade-offs. These results all together demonstrate Famba-V as a promising efficiency enhancement technique for Vim models.



%


\end{abstract}
\section{Introduction}
\label{sec:intro}

Transformer~\cite{vaswani2017attention, wan2023efficient, tao2024scaling, wan2024d2o, wan2024look}, which leverages the attention mechanism to excel in global perception, has become the dominant architecture across various vision tasks. However, the quadratic complexity of the attention mechanism with respect to sequence length makes Transformer-based methods less efficient.

In recent years, State Space Models (SSMs)~\cite{gu2023mamba, gu2021efficiently, mehta2023long, wang2023selective} have emerged as an alternative to Transformer-based methods with linear time complexity. Among SSMs, Mamba~\cite{gu2023mamba} is a representative work that integrates time-varying parameters into the SSM.
Due to its potential scaling and adaptability, Mamba opens up new possibilities for applying SSMs in vision tasks.
However, despite pioneering works such as Vision Mamba (Vim) \cite{zhu2024vision} adopting Mamba for vision tasks, the increased computational and memory demands during training can pose challenges. 
To achieve efficient training, techniques such as token pruning~\cite{rao2021dynamicvit, yin2022vit, meng2022adavit, liang2022not, kong2022spvit} and token fusion~\cite{bolyatoken, zhang2024mg, cao2023pumer, wan2024look, wan2024d2o} have been successfully adopted in the context of Transformer-based architectures such as Vision Transformer (ViT)~\cite{dosovitskiy2020image}. However, the exploration of token fusion techniques in Mamba-based architectures remains largely unexplored.

In this work, we introduce Famba-V, a simple yet effective technique based on cross-layer token fusion, with the objective to improve the training efficiency of Vision Mamba (Vim)~\cite{zhu2024vision}.
%
Similar to \cite{bolyatoken}, Famba-V measures the similarity between tokens to identify those containing similar information and fuse them together. 
More importantly, instead of blindly applying token fusion across all the layers within the Vim models, which as we will show, leads to a significant accuracy drop, Famba-V incorporates a suit of cross-layer token fusion strategies to select a subset of layers within the Vim models where token fusion should be performed.

We explore the effectiveness of Famba-V on CIFAR-100 dataset~\cite{krizhevsky2009learning}. 
Compared to vanilla Vim models without token fusion, Famba-V is able to reduce both training time and peak memory usage during training, demonstrating its effectiveness in enhancing the training efficiency of Vim models. 
Compared to the all-layer token fusion approach that existing works adopt, the cross-layer token fusion strategies incorporated by Famba-V offer improved accuracy-efficiency trade-offs.
Lastly, we have conducted experiments to understand the impact of the key hyperparameters that Famba-V incorporates on the accuracy-efficiency trade-offs.

In summary, our main contributions are listed as follows:
\vspace{-1mm}
\begin{enumerate}
    \item We propose Famba-V, a cross-layer token fusion technique that enhances the training efficiency of Vim models.
    \item Famba-V incorporates three cross-layer token fusion strategies to offer a better trade-off between accuracy and efficiency on Vim models.
    \item We conduct experiments on CIFAR-100. Our results show the effectiveness of Famba-V in enhancing the training efficiency of Vim models.
 
\end{enumerate}
\section{Related Work}

\subsubsection{State Space Models.} State Space Models (SSMs)~\cite{gu2021efficiently, gu2021combining, gupta2022diagonal} have emerged as a promising alternative to Transformers.
%
Gu et al.~\cite{gu2023mamba} introduce Mamba, which integrates time-varying parameters into the SSM and presents a hardware-aware algorithm for efficient training and inference. The superior scaling performance of Mamba shows it is a promising alternative to Transformers in language modeling.
Zhu et al.~\cite{zhu2024vision} extend the applicability of Mamba to vision tasks by proposing Vision Mamba (Vim), which captures long-range temporal dependencies in image data. Despite their potential, training these models remains challenging due to high computational and memory requirements. 

\vspace{-3.5mm}
\subsubsection{Efficient Mamba.} Recent research has focused on developing more efficient Mamba models for both natural language processing and computer vision. For language tasks, Ren et al.~\cite{ren2024samba} integrate Mamba with sliding window attention to efficiently model sequences with infinite context length. 
In some attempts to accelerate Mamba in visual representation, Lei et al.~\cite{lei2024dvmsr} introduce a lightweight image super-resolution network that leverages Vim and distillation to achieve efficient inference. 
Yao et al.~\cite{yao2024spectralmamba} propose an SSM framework, emphasizing spatial-spectral dynamics and computational downsizing without sacrificing accuracy. 
Pei et al.~\cite{pei2024efficientvmamba} explore the integration of visual SSMs and convolutions to balance between local and global extraction, demonstrating improved efficiency in vision tasks. 
Qin et al.~\cite{qin2024mambavc} introduce an SSM-based compression network optimized for visual data. 
Our work contributes to this growing body of research by focusing on token fusion for Vim models, offering a unique approach to enhance efficiency.

\vspace{-3.5mm}
\subsubsection{Efficient Vision Transformers.} 
Our work is also related to efficient vision transformers (ViTs). Efforts to create more efficient ViTs have followed various strategies. One approach focuses on pruning redundant neurons and connections \cite{zhu2021vision, tang2022patch, yang2023global}, while another aims to modify weights and activations to reduce memory usage \cite{yuan2022ptq4vit, ding2022towards, liu2023noisyquant}. Unlike token pruning, which removes tokens, token merging~\cite{bolyatoken} instead combines tokens, which aims to preserve visual information while accelerating ViT models. Existing methods like~\cite{zhang2024mg, cao2023pumer} have explored token merging in vision tasks. Building on these advancements, our work explores several cross-layer strategies for token fusion to optimize the accuracy-efficiency trade-off in Vision Mamba models. 
\section{Method}

\subsection{Preliminaries}
\subsubsection{Mamba.}
State Space Models (SSMs)~\cite{gu2021efficiently} map an input sequence $x(t) \in \mathbb{R}$ to an output sequence $y(t) \in \mathbb{R}$ through a hidden state $h(t) \in \mathbb{R}^\mathtt{N}$. In the 1-D situation, discrete SSM transform sequences as linear ordinary differential equations (ODEs):
\begin{equation}
\begin{aligned}
\label{eq:lti}
h'(t) &= \mathbf{A}h(t) + \mathbf{B}x(t), \\
y(t) &= \mathbf{C}h(t).
\end{aligned}
\end{equation}
Where $\mathbf{A} \in \mathbb{R}^{\mathtt{N} \times \mathtt{N}}$ is state transition matrix, while $\mathbf{B} \in \mathbb{R}^{\mathtt{N} \times 1}$ is input coefficient matrix and $\mathbf{C} \in \mathbb{R}^{1 \times \mathtt{N}}$ serves as an output matrix. 

Mamba~\cite{gu2023mamba, dao2024transformers} is an SSM-based model with selective state spaces. To improve expressiveness and flexibility, Mamba proposes to make $\mathbf{A}$ and $\mathbf{B}$ dynamically dependent on inputs, enabling an input-aware selective mechanism for better state-space modeling. Mamba approximates this ODE by discretizing $\mathbf{A}$ and $\mathbf{B}$ with a time step parameter $\mathbf{\Delta}$ using a zero-order hold trick:
\begin{equation}
\begin{aligned}
\label{eq:zoh}
\mathbf{\overline{A}} &= \exp{(\mathbf{\Delta}\mathbf{A})}, \\
\mathbf{\overline{B}} &= (\mathbf{\Delta} \mathbf{A})^{-1}(\exp{(\mathbf{\Delta} \mathbf{A})} - \mathbf{I}) \cdot \mathbf{\Delta} \mathbf{B}.
\end{aligned}
\end{equation}
After discretization, equation (\ref{eq:lti}) is reformulated as follows:
\begin{equation}
\begin{aligned}
\label{eq:discrete_lti}
h_t &= \mathbf{\overline{A}}h_{t-1} + \mathbf{\overline{B}}x_{t}, \\
y_t &= \mathbf{C}h_t.
\end{aligned}
\end{equation}

\subsubsection{Vision Mamba (Vim).}
The original Mamba model is designed for processing 1-D sequences. To adapt it for vision tasks, Vision Mamba (Vim)~\cite{zhu2024vision} reshapes a 2-D image, represented as $\mathbf{t} \in \mathbb{R}^{\mathtt{H} \times \mathtt{W} \times \mathtt{C}}$, into flattened 2-D patches, denoted as $\mathbf{x_p} \in \mathbb{R}^{\mathtt{J} \times (\mathtt{P}^2 \cdot  \mathtt{C})}$. In this context, $(\mathtt{H}, \mathtt{W})$ represents the dimensions of the input image, $\mathtt{C}$ is the number of channels, and $\mathtt{P}$ is the size of image patches. These patches $\mathbf{x_p}$ are then linearly projected to vectors of size $\mathtt{D}$ and position embeddings $\mathbf{E}_{pos} \in \mathbb{R}^{(\mathtt{J}+1) \times \mathtt{D}}$ are added as follows:

\begin{equation}
\begin{aligned}
\label{eq:embed}
\mathbf{T}_0 &= [\mathbf{t}_{cls};\mathbf{t}_p^1\mathbf{W};\mathbf{t}_p^2\mathbf{W};\cdots;\mathbf{t}_p^{\mathtt{J}}\mathbf{W}] + \mathbf{E}_{pos}, \\
\end{aligned}
\end{equation}
where $\mathbf{t}_p^{\mathtt{j}}$ refers to the $\mathtt{j}$-th patch of the image $\mathbf{t}$, and $\mathbf{W} \in \mathbb{R}^{(\mathtt{P}^2 \cdot \mathtt{C}) \times \mathtt{D}}$ is a learnable projection matrix. The class token, denoted as $\mathbf{t}_{cls}$, is used to represent the entire patch sequence. This token sequence ($\mathbf{T}_{\mathtt{l}-1}$) is then fed into the $\mathtt{l}$-th layer of the Vision Mamba encoder to generate output $\mathbf{T}_{\mathtt{l}}$. Finally, the output class token $\mathbf{T}_{\mathtt{L}}^0$ is normalized and passed through a multi-layer perceptron (MLP) head to generate the final prediction $\hat{p}$, as follows:
\begin{equation}
\begin{aligned}
\mathbf{T}_l &= \mathbf{Vim}(\mathbf{T}_{l-1}) + \mathbf{T}_{l-1}, \\
\mathbf{f} &= \mathbf{Norm}(\mathbf{T}_L^0), \\
\hat{p} &= \mathbf{MLP}(\mathbf{f}),
\end{aligned}
\end{equation}
where $\mathbf{Vim}$ represents the Vision Mamba layer, $\mathtt{L}$ is the total number of layers, and $\mathbf{Norm}$ denotes the normalization layer.

\subsection{Famba-V}
\begin{figure}[t]
    \centering
    \begin{subfigure}[b]{\textwidth}
        \centering
        \includegraphics[width=1.0\textwidth]{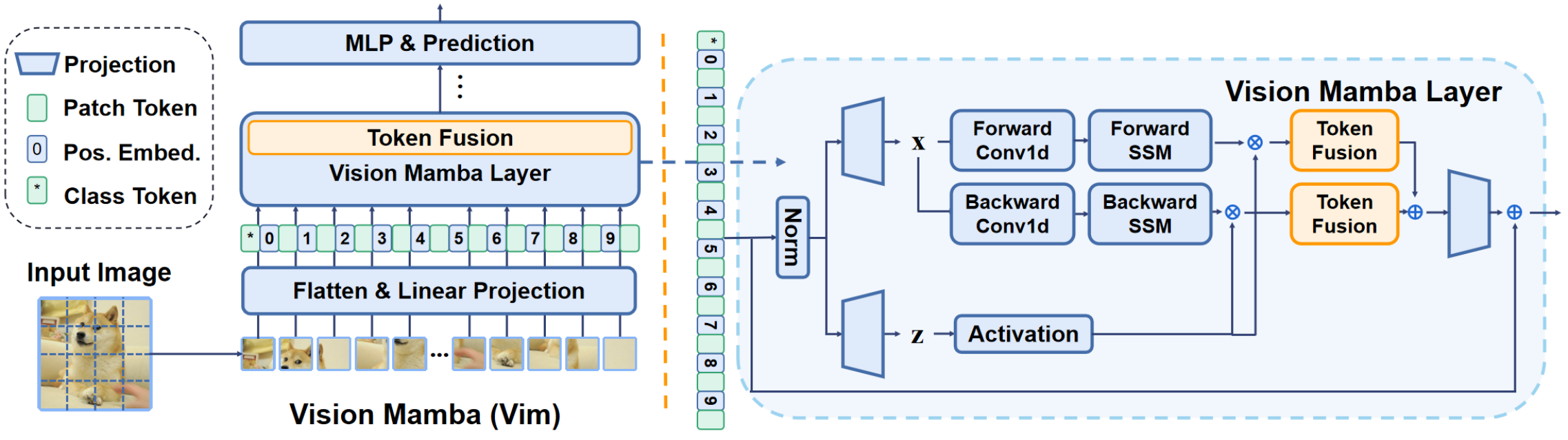}
        \caption{Overview of Famba-V.}
        \label{fig:fambav_overview_top}
        \vspace{2mm}
    \end{subfigure}
    \vspace{2mm}
    \begin{subfigure}[b]{\textwidth}
        \centering
        \includegraphics[width=1.0\textwidth]{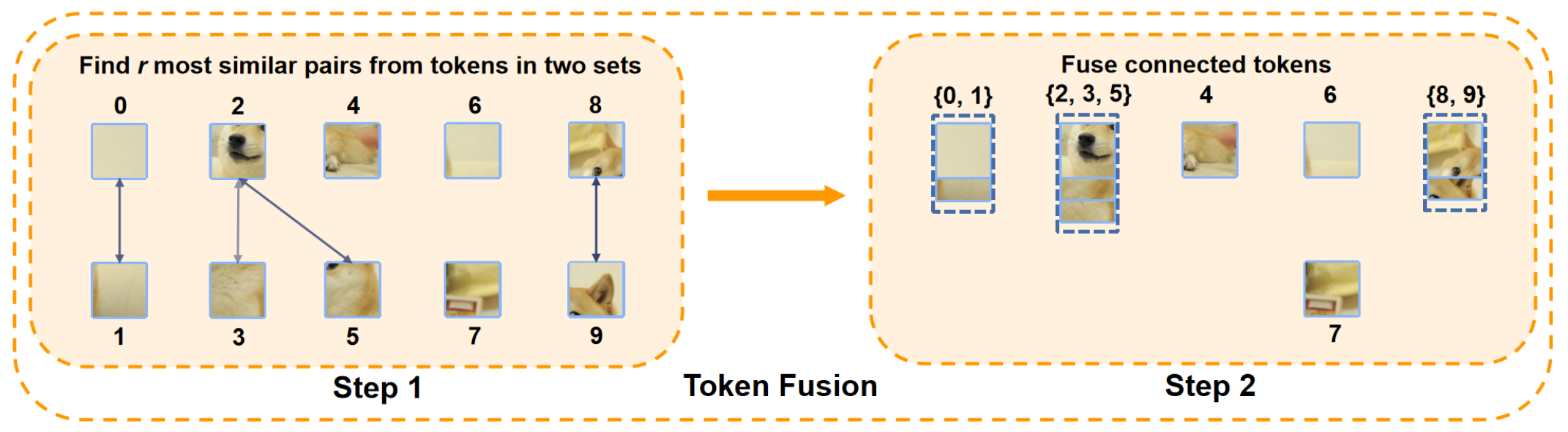}
        \caption{Illustration of token fusion.}
        \label{fig:fambav_overview_bottom}
    \end{subfigure}
    \caption{Overview of Famba-V and its token fusion design.}
    \vspace{6mm}
    \label{fig:fambav_overview}
\end{figure}

At a high level, Famba-V is a cross-layer token fusion-based method to enhance the training efficiency of Vision Mamba (Vim) models. 
Figure~\ref{fig:fambav_overview_top} provides an overview of Famba-V. 
Following Vim~\cite{zhu2024vision}, Famba-V splits the input image into patches and then projects them into patch tokens. The token sequence is concatenated with a class token as the input to the Vim layer. In Famba-V, the class token is designed to be placed at the head of the sequence so as to prevent it from being fused with other tokens by our proposed token fusion scheme. 

Inside the Vim layer, Famba-V applies token fusion to both the forward SSM and the backward SSM.
Figure~\ref{fig:fambav_overview_bottom} illustrates how token fusion is performed in Famba-V. Specifically, similar to~\cite{bolyatoken}, Famba-V first divides the input token sequence into two sets according to the even and odd indices of the input token sequence. It then utilizes cosine similarity to measure the similarity between tokens inside these two sets and identifies \textit{r} pairs of most similar tokens. Lastly, the paired tokens are fused together via averaging as a single token. 

\begin{figure}[t]
    \centering
    \includegraphics[width=1.0\textwidth]{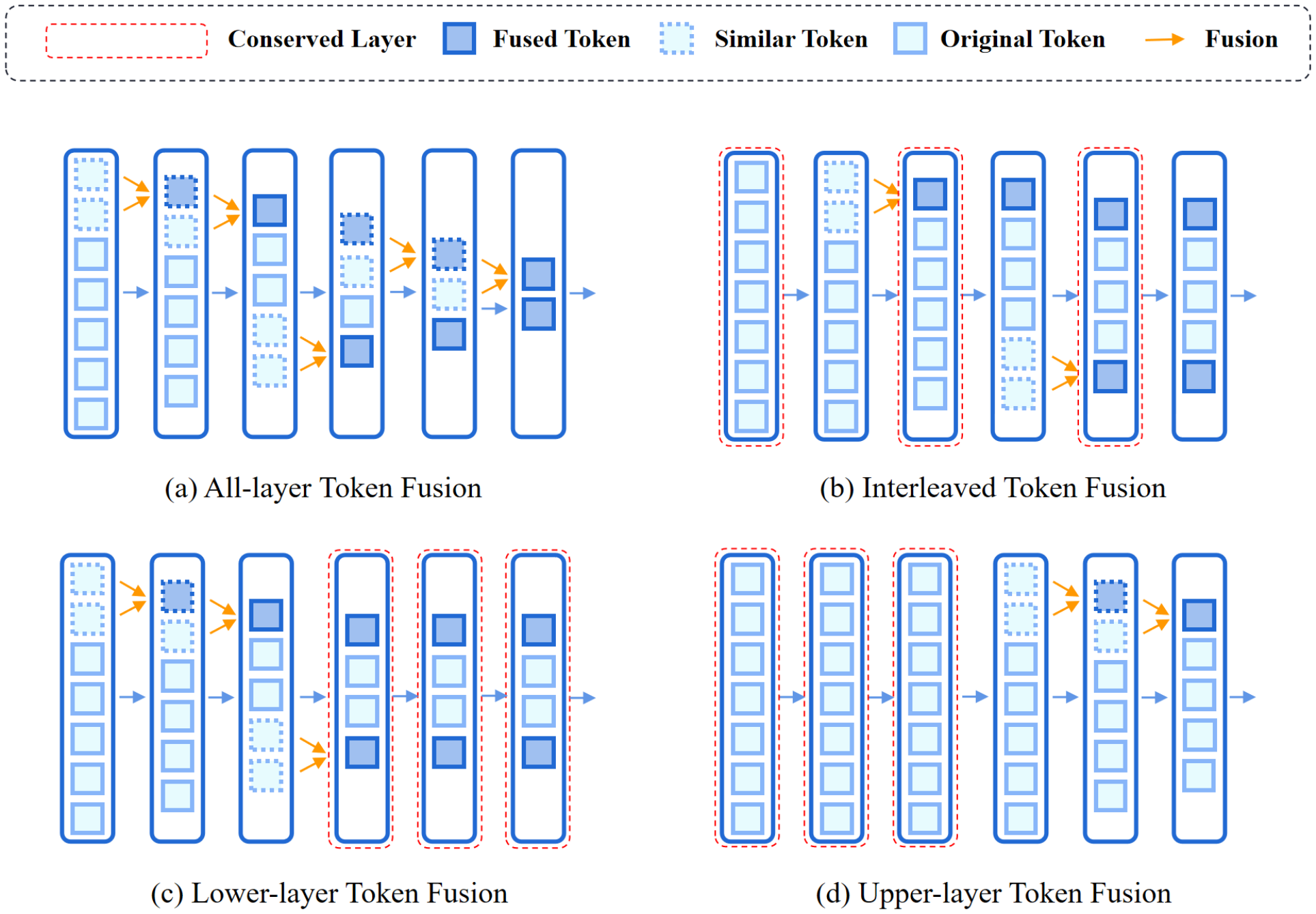}
    \vspace{-1mm}
    \caption{Illustration of (a) All-layer Token Fusion and three cross-layer token fusion strategies of Famba-V: (b) Interleaved Token Fusion, (c) Lower-layer Token Fusion, and (d) Upper-layer Token Fusion.}
    \label{fig:fusion_strategies}
    \vspace{-3mm}
\end{figure}

Figure~\ref{fig:fambav_overview_bottom} illustrates how token fusion is conducted in a single Vim layer. However, Vim models consist of multiple layers. The choice of layers for incorporating token fusion plays a key role in the trade-off between accuracy and efficiency of Vim models.
To explore this trade-off, as shown in Figure~\ref{fig:fusion_strategies}, Famba-V incorporates three different cross-layer token fusion strategies as follows. 
We initially apply token fusion to all layers of the Vim model, establishing this approach as our baseline strategy.

\vspace{-4mm}
\subsubsection{Strategy\#1: Interleaved Token Fusion.}
The interleaved strategy applies token fusion to alternate layers starting from the second layer.
This strategy is proposed to mitigate the potential negative impacts of aggressive token fusion while still reaping the efficiency benefits.

\vspace{-3mm}
\subsubsection{Strategy\#2: Lower-layer Token Fusion.}
The lower-layer token fusion strategy applies token fusion only to the lower layers while leaving the upper layers untouched.
This strategy is driven by the idea that early token fusion could lead to cascading efficiency gains throughout the network while preserving the model's higher-level reasoning capabilities.

\vspace{-3mm}
\subsubsection{Strategy\#3: Upper-layer Token Fusion.}
Lastly, the upper-layer token fusion strategy applies token fusion only to the upper layers while leaving the lower layers untouched.
This strategy is driven by the idea that higher-level features might be more amenable to fusing without significantly impacting the model's performance.

\vspace{2mm}
\noindent
It is important to note that for a fair comparison, we keep the total number of reduced tokens in these four strategies approximately the same. Additionally, for simplicity, for each of these four strategies, we keep the number of reduced tokens at each layer the same.

\section{Experiments}
\label{sec:exp}

\vspace{-1mm}
\subsubsection{Models and Datasets.}
We evaluate the performance of Famba-V using two variants of the Vim model architecture: Vim-Ti and Vim-S on CIFAR-100 \cite{krizhevsky2009learning}. Both Vim-Ti and Vim-S have 24 layers, with the key difference being their parameter counts: Vim-Ti has 7 million parameters, while Vim-S has 26 million. This difference arises from the hidden state dimension $D$ and expanded state dimensions $E$, which are set to $D = 192$, $E = 384$ for Vim-Ti, and $D = 384$, $E = 768$ for Vim-S.
 


\vspace{-4mm}
\subsubsection{Baselines and Evaluation Metrics.}
We compare Famba-V against two baselines: one using all-layer token fusion, and the other without any token fusion. 
We compare their performance under four metrics: top-1 accuracy, top-5 accuracy, training time, and the peak memory usage during training.

\vspace{-4mm}
\subsubsection{Implementation Details.}
We conduct our experiments on NVIDIA A100 (80GB) GPUs. We followed the training settings of Vim \cite{zhu2024vision} for a comparable analysis. Specifically, we applied data augmentations including random cropping, random horizontal flipping, label-smoothing regularization, mixup, and random erasing. We use AdamW with a momentum of 0.9, a batch size of 128, and a weight decay of 0.1. The training takes 300 epochs using a cosine learning rate schedule starting from 1 $\times$ $10^{-3}$ with the exponential moving average.

\begin{table}[t]
\centering
\caption{Comparison between baselines and Famba-V under three cross-layer token fusion strategies with Vim-Ti and Vim-S models on CIFAR-100.}
\vspace{-2mm}
\addtolength{\tabcolsep}{-1pt}
\begin{tabular}{>{\centering\arraybackslash}m{50pt} | >{\centering\arraybackslash}m{110pt} | >{\centering\arraybackslash}m{50pt} | >{\centering\arraybackslash}m{50pt} | >{\centering\arraybackslash}m{40pt} | >{\centering\arraybackslash}m{40pt}}
\toprule
Backbone & Strategy & Top-1 Accuracy  & Top-5 Accuracy & Training Time & Peak Memory \\
\midrule
Vim-Ti & w/o Token Fusion & 70.1 & 90.6 & 4:33:02 & 8302   \\
Vim-Ti & All-layer Token Fusion & 61.9 & 85.6 & 4:06:45 & 5499   \\
Vim-Ti & Interleaved Token Fusion & 67.0 & 89.3 & 3:59:21 & 5354   \\
Vim-Ti & Lower-layer Token Fusion & 62.2 & 86.1 & 3:55:56 & 4741   \\
Vim-Ti & Upper-layer Token Fusion & 70.5 & 90.5 & 4:08:40 & 5996 \\
\midrule
Vim-S & w/o Token Fusion & 76.9 & 93.6 & 7:46:16 & 16439   \\
Vim-S & All-layer Token Fusion & 68.9 & 89.3 & 6:01:16 & 10657   \\
Vim-S & Interleaved Token Fusion & 72.8 & 91.2 & 5:56:15 & 10400   \\
Vim-S & Lower-layer Token Fusion & 69.0 & 89.0 & 5:34:36 & 9139   \\
Vim-S & Upper-layer Token Fusion & 75.2 & 92.0 & 6:19:28 & 12161   \\
\bottomrule
\end{tabular}
\vspace{3mm}
\label{tab:cifar100maintable}
\end{table}

    
    

\subsection{Main Results}
First, we compare the performance of Famba-V under the three proposed cross-layer token fusion strategies against the two baselines on Vim-Ti and Vim-S models on CIFAR-100. 
When comparing with the all-layer token fusion baseline, for a fair comparison, we keep the total number of reduced tokens approximately the same across all four strategies. Specifically, for all-layer token fusion strategy, the total number of reduced tokens is 168, with 7 tokens reduced per layer; for interleaved token fusion strategy, the total number of reduced tokens is 168, with 14 tokens reduced per layer; for lower-layer token fusion strategy, the total number of reduced tokens is 171, with 9 tokens reduced per layer, starting from the 1st layer to the 19th layer; for upper-layer token fusion strategy, the total number of reduced tokens is 171, with 9 tokens reduced per layer, starting from the 6th layer.
%
Table~\ref{tab:cifar100maintable} summarizes our results. We have four observations. 
(1) In comparison with vanilla Vim-Ti and Vim-S without token fusion, Famba-V reduces both training time and peak memory usage during training under all three cross-layer token fusion strategies. This result demonstrates the effectiveness of token fusion in enhancing the training efficiency of Vim models. 
(2) For the all-layer token fusion baseline, although it achieves quite decent efficiency gains in terms of both training time and peak memory usage, it suffers from a large accuracy drop compared to the vanilla model without token fusion. 
%
(3) Among the three cross-layer token fusion strategies incorporated in Famba-V, the lower-layer token fusion strategy achieves the lowest accuracy, but provides the most significant reductions in training time and peak memory usage, making it a viable option for scenarios under resource constraints.
(4) In contrast, the upper-layer token fusion strategy preserves the accuracy to the greatest extent compared to the vanilla model without token fusion while still achieving decent efficiency gains.
%

%
Given the superiority of the upper-layer token fusion strategy, we use it as the default cross-layer token fusion strategy in the following experiments.

\begin{table}[t]
\centering
\caption{Impact of the selection of the starting layer to perform
token fusion on the training performance under the upper-layer token fusion
strategy on Vim-Ti.}
\addtolength{\tabcolsep}{-1pt}
\begin{tabular}{>{\centering\arraybackslash}m{71pt} | >{\centering\arraybackslash}m{40pt} | >{\centering\arraybackslash}m{55pt} | >{\centering\arraybackslash}m{55pt} | >{\centering\arraybackslash}m{55pt} | >{\centering\arraybackslash}m{40pt}}
\toprule
\# of Reduced Tokens per Layer & Starting Layer & Top-1 Accuracy  & Top-5 Accuracy & Training Time & Peak Memory \\
\midrule
15 & 15th & 72.7 & 92.2 & 4:22:04 & 7185  \\
14 & 14th & 73.6 & 92.2 & 4:20:34 & 7049  \\
13 & 13th & 72.7 & 92.2 & 4:19:23 & 6936   \\
12 & 12th & 72.3 & 92.0 & 4:15:32 & 6828   \\
12 & 11th & 72.0 & 91.6 & 4:14:33 & 6574   \\
11 & 10th & 71.9 & 91.5 & 4:12:54 & 6496   \\
10 & 9th & 71.7 & 91.1 & 4:12:35 & 6498   \\
10 & 8th & 70.6 & 90.9 & 4:10:20 & 6237   \\
9 & 7th & 71.0 & 91.2 & 4:13:21 & 6258   \\
9 & 6th & 70.5 & 90.5 & 4:08:40 & 5996   \\
8 & 5th & 70.5 & 90.5 & 4:09:31 & 6071   \\
8 & 4th & 69.6 & 90.2 & 4:11:19 & 5810   \\
7 & 3th & 68.1 & 89.5 & 4:08:11 & 6007   \\
7 & 2th & 66.8 & 88.9 & 4:16:27 & 5764   \\

\bottomrule
\end{tabular}
\label{tab:cifar100layerselection}
\end{table}

    
    

\subsection{Impact of the Selection of Starting Layer}
Next, we examine the impact of the selection of the starting layer to perform token fusion on the training performance under the upper-layer token fusion strategy on Vim-Ti on CIFAR-100. 
Starting layer refers to the first layer in the model from which token fusion is performed. For example, if the starting layer is the 8th layer, it means that token fusion is added from the 8th layer of the model and continues through the subsequent layers. 
%
Table~\ref{tab:cifar100layerselection} summarizes our results when changing the starting layer from 2nd to 15th. For a fair comparison, we keep the total number of reduced tokens across all the starting layer scenarios approximately the same. Therefore, the number of reduced tokens per layer is different under different starting layer scenarios. 
Note that we only show the results from the 2nd to the 15th layers because starting from the 1st is equal to the all-layer token fusion strategy, while starting from the 15th layer and beyond has limited efficiency gain.
%
As shown, the relationship between efficiency and the starting layer is not linear, as the highest efficiency is achieved when the starting layer is 6th. We conjecture that for scenarios where the starting layer is lower than the 6th layer, even though token fusion is performed at an early stage, the number of reduces tokens per layer is small such that the efficiency gain is limited; for scenarios where the starting layer is higher than the 6th layer, although the number of reduced tokens per layer is larger, token fusion is performed at a later stage, which hurts its efficiency benefits.


\subsection{Impact of the Number of Reduced Tokens per Layer}
Lastly, we examine the impact of the number of reduced tokens per layer on the training performance under the upper-layer token fusion strategy on Vim-Ti on CIFAR-100.
For a fair comparison, we fix the starting layer at the 6th layer given that it achieves the highest efficiency as shown in Table~\ref{tab:cifar100layerselection}, and change the number of reduced tokens per layer from 1 to 9.
%
Table~\ref{tab:cifar100tokenreduction}
summarizes the results. 
%
As shown, there is a trade-off between accuracy and efficiency. Specifically, as the number of reduced tokens at each layer increases from 1 to 9, we observe a consistent reduction in training time and peak memory usage while accuracy drops as a trend. 

\begin{table}[t]
\centering
\caption{Impact of the number of reduced tokens per layer on the training performance under the upper-layer token fusion strategy on Vim-Ti.}
\addtolength{\tabcolsep}{-1pt}
\begin{tabular}{>{\centering\arraybackslash}m{71pt} | >{\centering\arraybackslash}m{40pt} | >{\centering\arraybackslash}m{55pt} | >{\centering\arraybackslash}m{55pt} | >{\centering\arraybackslash}m{55pt} | >{\centering\arraybackslash}m{40pt}}
\toprule
\# of Reduced Tokens per Layer & Starting Layer & Top-1 Accuracy  & Top-5 Accuracy & Training Time & Peak Memory \\
\midrule
9 & 6th & 70.5 & 90.5 & 4:08:40 & 5996  \\
8 & 6th & 70.8 & 90.4 & 4:11:57 & 6313   \\
7 & 6th & 70.4 & 90.7 & 4:15:31 & 6662   \\
6 & 6th & 70.8 & 91.2 & 4:20:17 & 6984   \\
5 & 6th & 70.9 & 91.1 & 4:24:59 & 7322   \\
4 & 6th & 70.9 & 91.2 & 4:34:12 & 7645   \\
3 & 6th & 71.7 & 91.4 & 4:38:33 & 7986   \\
2 & 6th & 71.4 & 91.6 & 4:42:38 & 8337   \\
1 & 6th & 72.4 & 91.6 & 4:42:32 & 8693   \\
\bottomrule
\end{tabular}
\vspace{3mm}
\label{tab:cifar100tokenreduction}
\end{table}
\vspace{-0mm}

\section{Conclusion}
In this paper, we presented Famba-V, a cross-layer token fusion technique that enhances the training efficiency of Vision Mamba models. 
Famba-V incorporates three cross-layer strategies to perform token fusion.
%
Our experimental results show that Famba-V reduces both training time and peak memory usage during training under all three cross-layer token fusion strategies and offer improved accuracy-efficiency trade-offs compared to the all-layer token fusion approach that existing works adopt. 
%
In our future work, we plan to further optimize the efficiency of Vision Mamba by combining token fusion with other efficiency enhancement techniques~\cite{wan2023efficient, wang2024svd}.





\section{Acknowledgement}

We thank the reviewers for their helpful comments. This work was partially supported by NSF under award NeTS-2312675.


%
%
\bibliographystyle{splncs04}
\bibliography{main}
\end{document}